\documentclass[sigconf, review=false]{acmart}

\usepackage{booktabs} % For formal tables
\renewcommand\footnotetextcopyrightpermission[1]{} % removes footnote with conference information in first column
\usepackage{xcolor}
\usepackage{multirow}
\usepackage{diagbox}
\usepackage{rotating}
\usepackage{tabulary}
\usepackage{colortbl}
\usepackage{tabularx}

\definecolor{RED}{rgb}{1,0,0}
\definecolor{BLUE}{rgb}{0,0,1}

\begin{document}
\title{Web Robot Detection in Academic Publishing}
% \titlenote{Produces the permission block, and
% \subtitle{Extended Abstract}
% \subtitlenote{The full version of the author's guide is available as
%   \texttt{acmart.pdf} document}

\author{Athanasios Lagopoulos}
\affiliation{%
  \institution{Aristotle University of Thessaloniki}
  \city{Thessaloniki} 
  \country{Greece}
}
\email{lathanag@csd.auth.gr}

\author{Grigorios Tsoumakas}
% \authornote{The secretary disavows any knowledge of this author's actions.}
\affiliation{
  \institution{Aristotle University of Thessaloniki}
  \city{Thessaloniki} 
  \country{Greece}
}
\email{greg@csd.auth.gr}

\author{Georgios Papadopoulos}
% \authornote{This author is the
%   one who did all the really hard work.}
\affiliation{%
  \institution{Atypon Systems Inc.}
  \city{Santa Clara}
  \state{California}
  \country{USA}}
\email{georgios@atypon.com}

% \affiliation{%
% \affiliation{%
% \affiliation{%
%   \country{China}
% }

% \affiliation{%


% \author{Julius P.~Kumquat}

% \renewcommand{\shortauthors}{B. Trovato et al.}

\begin{abstract}
Recent industry reports assure the rise of web robots which comprise more than half of the total web traffic. They not only threaten the security, privacy and efficiency of the web but they also distort analytics and metrics, doubting the veracity of the information being promoted. In the academic publishing domain, this can cause articles to be faulty presented as prominent and influential. In this paper, we present our approach on detecting web robots in academic publishing websites. We use different supervised learning algorithms with a variety of characteristics deriving from both the log files of the server and the content served by the website. Our approach relies on the assumption that human users will be interested in specific domains or articles, while web robots crawl a web library incoherently. We experiment with features adopted in previous studies with the addition of novel semantic characteristics which derive after performing a semantic analysis using the Latent Dirichlet Allocation (LDA) algorithm. Our real-world case study shows promising results, pinpointing the significance of semantic features in the web robot detection problem.
\end{abstract}


% <concept_id>10002951.10003260.10003277.10003280</concept_id>
% <concept_desc>Information systems~Web log analysis</concept_desc>
% <concept_id>10010147.10010257.10010258.10010259.10010263</concept_id>
% <concept_desc>Computing methodologies~Supervised learning by classification</concept_desc>
% <concept_significance>300</concept_significance>
% <concept_id>10002951.10003317.10003318.10003320</concept_id>
% <concept_desc>Information systems~Document topic models</concept_desc>
% <concept_significance>100</concept_significance>

% \ccsdesc[500]{Information systems~Web log analysis}
% \ccsdesc[300]{Computing methodologies~Supervised learning by classification}
% \ccsdesc[100]{Information systems~Document topic models}


\maketitle

\section{Introduction}

Web (ro)bots, also known as web crawlers, are computer programs that request resources from web servers across the Internet without human intervention. The constant growth of Web 3.0 technologies and social media generate a huge amount of valuable information ready to be accessed by both traditional web crawlers and emerging advanced robots representing Internet of Things devices, such as smart watches, cars and digital assistants \cite{rude2015request}. As of 2016, the web traffic originated by web bots constitutes more than a half (51.8\%) of the total web traffic, being in an uptrend after three years of decline \cite{zeifman2016}.

{\em Malicious} bots threaten the security, privacy and performance of a web application. {\em Non-malicious} bots are involved in analytics skewing, affecting the reliability of metrics and, by extension, the decision making process \cite{doran2011web}. A recent industry report \cite{distil2017} points out that large websites with unique content, such as blogs, on-line newspapers and digital libraries of academic publications, are the most attractive to bots. The most common threat that such websites need to deflect is skewing: their metrics and ratings are altered, intentionally by malicious robots and unintentionally by non-malicious robots, rendering their validity questionable and giving the false impression that some piece of information is highly popular and recommended by many \citep{greene2016web}. In addition to this, social bots contribute further to the spread of unverified information or rumors. Therefore, the detection of web robots and by extension the filtering out of their activities are the actions need to be taken so as to maintain the higher resolution picture of unbiased and unaltered information shared in the web \citep{ferrara2016rise}.

% Several types of web robots have been identified based on their behavioral characteristics and functionality. Malicious bots threaten the security, privacy and performance of a web application. {\em Scrappers}, for example, extract content and duplicate it, steal unique data and hurt search engine optimization \citep{lee2009classification}. {\em Impersonators} bypass security solutions and initiate distributed denial-of-service attacks \citep{wang2015ddos}. {\em Hacker tools}, which are very common for websites with sign up/in pages, can create fake accounts and crack user credentials and can even lead to credit card fraud. 

% Movie, music and other recommendation websites are affected by fake user ratings which harm their suggestion algorithms and methods.

% In the scholarly and publishing domain, the emergence of altmetrics \citep{priem2010altmetrics} got the attention of web robots . These metrics are altered, intentionally by malicious robots and unintentionally by non-malicious robots,  

%These techniques can be divided into four types: {\em syntactical or heuristic log analysis}, {\em traffic pattern analysis}, {\em analytical learning techniques} and {\em Turing test systems}. While the first three techniques are used in off-line detection, the last one is used in real-time. However, recent studies have also used analytical techniques in real-time detection of web robots. \cite{doran2016integrated}, \cite{wang2015ddos}. %

%These techniques can be further classified according to the learning algorithms and the features they adopt.

This paper introduces a novel web robot detection approach for content rich websites. The key assumption of the proposed approach is that humans are typically interested in specific topics, subjects or domains, while robots typically crawl all the available resources irrespectively of their content, with the exception of a special class of web robots, called focused crawlers \cite{diligenti2000focused}. Based on this assumption, our main contribution is a novel class of features, for representing user sessions, that capture the semantics of the content of requested resources. Correspondingly, our main research question is whether such features can improve the results of supervised learning approaches to web robot detection in content rich websites. Empirical results on real-world data from the digital library of an academic publisher provide evidence in favor of a positive answer to our research question. 

%a when we enrich  traditional representations of user sessions with such features, we achieve improved web robot detection results.

%In particular, we apply the Latent Dirichlet Allocation (LDA) algorithm to the text of the resources 

%extraction of semantic features from the content of the requested resources 

%on web robot detection which takes advantage of the content web applications serve. We use a standard supervised learning method to train a bot detection model and we focus on the enrichment of previously used features with novel semantic features extracted from the content of requested resources or pages using the LDA algorithm. 

This paper is structured as follows: after providing the background and related work in Section \ref{relate} , we introduce our approach on extracting semantics from sessions in Section \ref{semantics}. In Section \ref{empirical}, we describe our real world case study by presenting the data and the steps taken before creating a detection model, while in Section \ref{results} we discuss the results of our study proving our assumption. Finally, in Section \ref{conclude}, we review our approach and draw some future directions. 

\section{Background and Related Work}
\label{relate}

\subsection{Web Robot Taxonomy}

Several categories of malicious web robots have been defined based on their behavioral characteristics and range of capabilities. {\em Scrappers}, for example, collect application content and data for use elsewhere and obtain limited-availability goods and services by unfair methods \cite{lee2009classification}. {\em Hacker tools} are involved in credit card, credential and token cracking and target resources of the application and database servers to achieve denial of service (DDOS), while {\em impersonators'} functionality spans from account creation and spamming to ad fraud via false clicks and sniping by performing last minute bids for goods \cite{wang2015ddos}. In addition, a taxonomy listing the automated threats on web applications \cite{owasp2016} has been created by the community of the Open Web Application Security Project\footnote{\url{http://www.owasp.org/}} (OWASP). 

Similarly, a variety of different categories of non-malicious bots have been defined. {\em Search engine crawlers} collect information to improve their ranking algorithms, {\em feed fetchers} carry the information of a website to a web or mobile application, while {\em monitoring} bots help developers keep track of the health and function of their website \citep{doran2013comparison}. Interestingly enough, feed fetchers comprise more than 12\% of the total traffic, with Facebook's mobile application feed fetcher being the most active web robot accounting for 4.4\% of all website traffic \cite{zeifman2016}. 

\subsection{Web Robot Detection Approaches}
There are four main categories of web robot detection approaches: {\em syntactic log analysis}, where string processing techniques are used; {\em analytical learning}, which make use of machine learning algorithms and features that contain different properties deriving from the user sessions in the server; {\em traffic pattern analysis}, which search for statistical diversity between the features of human and robots; {\em Turing test systems}, which identify a robot in real-time by means of a Turing test. Recent work on web robot detection mainly focuses on analytical learning approaches, which achieve considerably better results than the other categories of approaches, since the latter rely on procedures or algorithms that a properly engineered bot can evade \citep{doran2011web}.   

\subsection{Analytical Learning}
\label{sec:analytical_learning}

An important first step in analytical learning approaches is {\em session identification}, which is concerned with breaking the click stream into sessions. Various timeout thresholds have been investigated in the past for session identification, such as 10 minutes \cite{alnoamany2013access}, 30 minutes \cite{tan2004discovery} and using dynamically adaptive thresholds ranging from 30 minutes to 60 minutes \cite{stassopoulou2007probabilistic}.

\begin{figure*}[bt]
\centering
\includegraphics[width=0.9\textwidth]{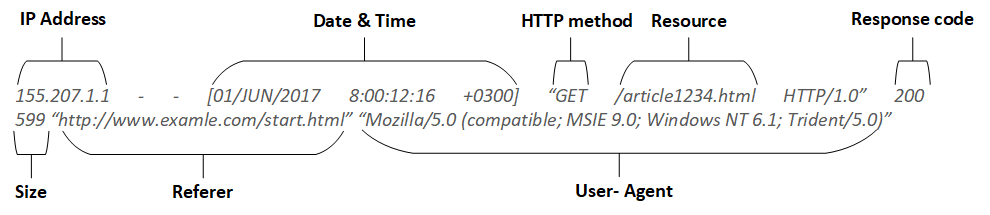}
\caption{Example of an entry in a web server access log file.}
\label{fig:request_simple}
\end{figure*}

An important second step concerns feature extraction from the identified sessions, based on the variety of information found in the entries of web server access log files, such as: the {\em IP address} of the host that made the request to the server, the {\em date and time} that the request was received, the {\em resource} requested, the {\em HTTP method} used (e.g GET, HEAD, POST), the {HTTP response code} sent back to the client (200, 404 etc.), the {\em size} of the returned object, the {\em Referer HTTP request header}, which is the page that links to the resource requested and the {\em User-Agent String} that identifies the client's browser. Figure \ref{fig:request_simple} shows an example of an entry in a server access log file.

Some of the typical features used by analytical learning approaches are: 
\begin{itemize}
\item {\em Total Requests}. The total number of requests in a session.
\item {\em Session Duration}. Total time, in seconds, elapsed between the first and the last request of a session. 
\item {\em Average Time}. Average time, in seconds, between two consecutive requests.
\item {\em Standard Deviation Time}. The standard deviation of time between two consecutive requests. This feature corresponds to the visitors browsing behavior. Users usually don't spend the same time in each page. A page of an interesting article will keep them, in the page, for a much longer time than just a reference page. On the other hand, bots usually keep a stable browsing pattern where they visit each page for almost the same amount of time. Therefore, the standard deviation of humans will be greater than that of robots. 
\item {\em Repeated Requests}. Percentage of repeated requests of a page in a session. As repeated request is counted when two identical requests occur in a session. Identical requests are considered those that visit the same page and with the same HTTP method.
\item {\em HTTP requests}. This feature consist of 4 different values. Each value represents the percentage of requests of the different HTTP response codes. We separate the HTTP response codes in 4 categories depending on the first digit of their code: Successful requests - 2xx, Redirection - 3xx, Client Errors - 4xx and Server Errors - 5xx.
\item {\em Specific Type Requests}. The percentage of type X requests over the number of all requests. This feature can be modified according to the application. 
\end{itemize}

Many websites are based on content management systems or specialized web applications that log additional information about the identity of the visitors of a website, beyond that found in web server access logs. Such information may be the country that the request is coming from, by checking the user's IP using a geolocation service; the username of a logged in user or an indicator if the request originates from a web service. This kind of information can be a great source of valuable features for a web robot detector, but unfortunately it normally is application dependent and not always available.

\subsection{Related Work}

Several supervised and unsupervised analytical learning approaches have been developed in the past based on a variety of learning algorithms and features. Tan et al.~\cite{tan2004discovery} used decision trees (C4.5 algorithm) to train a model using 25 different features that were extracted from each user session. The feature vector included percentages of the different content (images, multimedia, HTML etc.), time characteristics (average  time, total time etc.), request types (GET, POST, HEAD etc.) and other (IP, user-agent etc.). Bomhardt et al.~\cite{bomhardt2005web} used neural networks and included features like total number of bytes and percentage of response codes (200, 2xx, 404 etc.). Stassopoulou et al.~\cite{stassopoulou2009web} used a heuristic semi-automatic method to label the training data and introduced a Bayesian approach to classify the sessions. Stevanovic et al~\cite{stevanovic2012feature} experimented with a variety of classifiers (C4.5, RIPPER, k-nearest, Naive Bayesian, Bayesian Network, SVM and Neural Networks) and introduced two novel features considering the page depth of a session's requests and the sequentiality of HTTP requests. Finally, the recent research of Doran et al.~\cite{doran2016integrated} presents a novel approach, that in contrast to the approaches mentioned earlier, can also be used for real-time detection of web robots. Their approach is based on a first-order discrete time Markov chain model and the request patterns of the visitors. Also, in order to embed the detection algorithm in a real-time system they present a new way to identify sessions from log file entries. In contrast with the above supervised approaches, the study of Stevanonic et al.~\cite{stevanovic2013detection} used unsupervised neural networks to detect humans and robots and to further analyze the behavior of malicious and non-malicious web robots, while Zabihi et al.~\cite{zabihi2014density} used the DBSCAN clustering algorithm with just four different features.

Few studies have addressed web robot detection in the particular domain of academic publishing, where our empirical study is focused. Huntington et al. \cite{huntington2008web} was the first to examine this domain, comparing the activity of robots in open access and restricted full text articles of a biology journal. Robots were identified using different heuristic methods and behavioral pattern techniques without using any machine learning. A second recent study benchmarked existing web robot detection approaches in Open Access institutional repositories \cite{greene2016web}. By performing a close review of the literature, system documentation and open source code, the study concludes that web robot detection is most successful when a variety of data and techniques are combined and pinpoints that non of the examined methods leads to usage statistics that are completely free of robot activity. 

\section{Extracting Semantics from Sessions}
\label{semantics}

As already mentioned in the introduction, this work is based on the assumption that humans typically look for specific information on a particular topic, while on the other hand, most of the web robots go through the content of a website in a uniform fashion, without favoring specific pages or content. Building a web robot detection approach on top of this assumption, requires measuring the semantic (in)coherence of the content visited during a session. 

To achieve this, we start with topic modeling of the content of a website using latent Dirichlet allocation (LDA) \cite{blei2003latent}. LDA describes each document or, in this case, each web resource, as a probability distribution over a user-defined number, $k$, of  topics, where each topic is a probability distribution over words. 

Consider a session, $S$, comprising $n$ requests for web pages (or other textual resources, such as PDF files). Let $p_ij$, be the probability of topic $j$, $1\leq j\leq k$, for the web page associated with request $i$, $1\leq i\leq n$. Let also $p_i$ be a vector containing the distribution over the $k$ topics for the web page associated with request $i$. We extract the following {\em semantic features} from $S$:

%We define a session, $S$, as a sequence of $N$ content resources $(r_1, r_2, \ldots, r_N)$ that were visited during the session.  where $p_i$ is a page visited by the user during the session and $t_{i,j}$ the $j$'th ($0<j\leq k$) topic of that page. As $P(t_{i,j})$ we set the probability of a topic $t$. 
%We then extract the following {\em semantic features} from a session, based on the probability distributions over topics of the visited pages within this session: 

\begin{itemize}
\item {\em Total Topics} (TT). The number of topics with non-zero probability.
$$
\text{TT} = |\{(i,j): 1\leq i\leq n, 1\leq j\leq k, p_{ij} \neq 0\}| % \sum_{i=1}^{N}\sum_{j=1}^{k} 1, \text{ if }      P(t_{i,j})\neq0
$$
The higher the total number of topics with non-zero probability in all requests of a session, the lower the semantic coherence of the session.%\GT{I would also try to divide this feature by $n$. Otherwise it is affected also by session length. Just a note for the future.}
\item {\em Unique Topics} (UT). The number of unique topics with non-zero probability in a session. 
$$
	\text{UT} = |\{j: 1\leq j\leq k, (\exists i) p_{ij} \neq 0\}|
%	\text{UT} = \sum_{j=1}^{k}e
$$
%Where $e=1$, if $P(t_{i,j})\neq 0$ and $p_i \in S$ and $e=0$, otherwise.
This feature also measures the semantic inconsistency of a session, but without counting the same topic twice. %\GT{I think this should also have been divided by $n$. Otherwise it is affected also by session length. Just a note for the future.}
\item {\em Page Similarity} (PS). The percentage of unique topics with non-zero probability over all the topics with non-zero probability of a session.
$$
	PS = \dfrac{UT}{TT}
$$
This feature models the dissimilarity of the different pages visited during a session. The lower its value, the more semantically similar the requested resources.
\item {\em Page Variance} (PV). The semantic variance of the pages of a session. 
$$
	PV = \dfrac{\sum_{i} \sqrt{\sum_{j=1}^{k}(p_{ij}-\overline{p}_j)^2}}{n},
$$
where $\overline{p}=\frac{1}{n}\sum_{i=1}^{n}p_i$ is the mean of the vectors $p_i$ associated with each request of the session. This feature computes the mean Euclidean distance of the topic distribution of the resource of each request with that of the mean topic distribution. The lower this distance, the higher the semantic similarity of the requested resources in the session.  
\item {\em Boolean Page Variance}. It is a boolean version of PV, where prior to its calculation we set all non-zero $p_{ij}$ values to be equal to 1.
% \item Page Variance (PV): The semantic variance of the pages of the session. The topics of each page are represented as Boolean vectors; the $i$'th component of the vector is $1$ if the page belongs to the topic $c_i$ and $0$ otherwise. The arithmetic mean of a set of such vectors contains as $i$'th component the proportion of pages of the set belonging to the topic $c_i$. We define the variance of a session, which contains a set of pages, as the average squared distance between each page's topic $t_i$ and the set's mean vector $\overline{t}$, i.e.,
% $$
% 	VP = \dfrac{\sum_{i} d(t_i,\overline{t})^2}{|N|},
% $$
% where $d(t_i,\overline{t})^2$ is the Euclidean distance
% $$
% 	d(t_1,t_2) = \sqrt[]{\sum_{i}(t_{1,i} - t_{2,i})^2},
% $$
% where $t_{k,i}$, it the $i$'th component of the topics vector of a page $x_k$. As $|N|$ we define the total number of the different pages found in a  session. 
% \item Probabilistic Variance: It is a variation of the {\em Page Variance} feature, where the topics of each page are not represented as Boolean vectors but as vectors which contain the probability of the topic; the $i$'th component of the vector is the probability of the topic $c_i$ of a page, if the page belongs to this topic.   
\end{itemize}
 
To the best of our knowledge, this is the first work to consider modeling the semantics of the content of web sites for detecting web robots. 
 
\section{Real World Case Study}
\label{empirical}
In this section, we present the real world case study where we applied the proposed approach. First, we present the dataset and its pre-processing procedure. Then, we discuss our session identification method and we define the features extracted for our detection model. Finally, we unfold our three-step labeling process. 
% Our pipeline is based in previous analytical learning approaches \citep{stassopoulou2007probabilistic,stevanovic2012feature} 

% Each of the above steps are described in the following sections.
% \SL{Maybe we don't need this}Figure \ref{fig:steps} shows an overview of the training and classification stages followed. 

% \begin{figure}[ht]
% \centering
% \includegraphics[width=\columnwidth]{figures/steps}
% \caption{The web robot detection pipeline.}
% \label{fig:steps}
% \end{figure}

\subsection{Dataset Preparation}
\label{dataset}
Our data come from the web portal of a large commercial academic publisher. The data concern log files of the server requests for an entire month (January 2014). The log files contain in total 25.318.451 requests without including those entries that don't contain sufficient length of text or any semantic value. These entries may include the home page, search results, log in page, about page, contact page, etc. The remaining entries include only the following requests regarding the available articles in the library:
\begin{itemize}
\item Abstract (HTML): The web page containing the abstract of an article.
\item Full-text (HTML): The web page containing the full-text of an article.
\item Full-text (PDF): The PDF file containing the full-text of an article.
\item References (HTML): The web page containing the references of an article.
\item Supplementary Material (HTML): The web page containing supplementary material (Tables, Data etc.) of an article. 
\end{itemize}

The aim of keeping pages that have these particular text content is to reduce the size of the log entries while keeping all the useful information that will enhance the semantically rich knowledge of which the semantic features proposed take advantage of.

Besides the log files we were also provided with the full corpus of the available articles in the digital library. In total, there are 2.253.533 different articles.

\subsection{Session Identification}
\label{session_ident}
In this step of our pipeline we identify distinctive user sessions. Log entries or requests are grouped together forming sessions. Generally, a session contains requests coming from a specific user, who is identified by both the IP address and the user-agent string. The user's requests break into different groups by applying a timeout threshold. In our case, the timeout threshold was set to 30 minutes following what appears to be the standard. Furthermore, it is reasonable to assume that any session with less than 3 requests is too short to produce significant features; thus, we ignore sessions with less than 3 requests in total.

The session identification step further reduces the amount of data by recognizing 10.039.241 sessions and the sessions that have more than two requests are just 1.727.568. The sessions seems to vary a lot for both the number of requests and the duration. They have an average (median) of 7.8 (4) requests while there are sessions with more than 10.000 requests. The average (median) duration of session is 629.5 (208.5) seconds and the average time-threshold between two consecutive requests is 131.6 seconds. The users of the sessions have in total visited 1.653.999 unique articles.

\subsection{Feature Extraction}

We extracted two sets of features from the identified sessions: {\em simple features}, based on past approaches of the literature, and {\em semantic features}, based the approach that we presented in this paper.

Simple features, include the 10 features discussed in Section \ref{sec:analytical_learning} ({\em HTTP requests is broken down into 4 features}), plus the following features that come from the particular web application of the publisher: 
\begin{itemize} 
\item {\em Unique Content}. Text content usually exists in different formats in a web page (pdf, eps, html etc.). This feature represents the number of the unique requests to this content regardless its format.
\item {\em Multiple Countries}. Indicating if the requests are coming from different countries ($>1$) during the sessions. The web application uses a geolocation service to accurately determine the location of the visitors using their IP address.
\item {\em Web Service}. Indicates whether the session comes from a web service or an application programming interface (API) of the web application or not. 
\end{itemize}

Note also that we customize the feature {\em Specific Type Requests} to {\em PDF requests}, measuring the  percentage of PDF requests over the number of all requests, so as to match our case study.

% Scholarly digital libraries, like the one this study works with, serve unique scientific papers that span across a variety of scientific fields. Thus, it makes       In our case, the use of semantic features, obtained from the articles visited, relies on the assumption that humans will be interest in specific topics or subjects while web robots may crawl resources unrelated to each other.

To extract the semantic features we first applied the LDA algorithm on the full corpus of the 2.253.533 available articles, since every page is associated with an article as specified in Section \ref{dataset}. The number of  topics was set to $k=5000$, but for each article only the top-10 topics with the highest probabilities were considered. Most of the articles (2.177.524 - $\approx96.6\%$) have non-zero probabilities in 10 topics and 6.446 ($\approx0.02\%$) articles have non-zero probability in just 1 topic. Finally, for each session, we constructed the five features proposed in Section \ref{semantics}.

% The histogram in Figure \ref{fig:topics_histogram} shows the distribution of the topics' probabilities in the log scale. As the number of buckets was set to 100, we notice that most articles belong to topics with probability less than 10\%, indicating weak connection between an article and its topics.

% \begin{figure}[ht]
% \centering
% \includegraphics[width=\columnwidth]{figures/histogram_log}
% \caption{A histogram showing the distribution of the topics' probabilities in the log scale.}
% \label{fig:topics_histogram}
% \end{figure}

\subsection{Session Labeling}
We identify and label a session as human or robot. The robot labeling procedure consists of three stages. During the first stage we label each session using the API from useragentstring.com\footnote{http://www.useragentstring.com/}. The API indicates if the user-agent string of a session belongs to one of the following categories: Cloud Client, Console, Offline Browser, Link Checker, Crawler, Feed Fetcher, Library, Mobile Browser, Validator, Browser, Unknown and Other. We consider as web robots only the user-agents labeled as Crawler.

In the second stage, we use two lists containing regular expressions that match with the user-agent string of known bots. The first one\footnote{https://github.com/atmire/COUNTER-Robots} is the official list of user agents that are regarded as robots/spiders by project COUNTER\footnote{https://www.projectcounter.org/}, which provides a code of practice that helps librarians and publishers record and report online resource usage stats in a consistent and credible way. The second list\footnote{https://goo.gl/5WQ6ds} is a regularly updated list that the open source web analytics software Piwik\footnote{https://piwik.org/} uses. The sessions whose user-agent string matches one of the regular expressions are labeled as robots. After sampling and manually checking some of the labeled sessions, we decided to remove the following questionable regular expressions of the first list: { \em \^{}Mozilla\$, \^{}Mozilla.4\textbackslash{}.0\textdollar{} } and  {\em \^{}Mozilla.5\textbackslash{}.0\textdollar{}}. Beside the sessions categorized as Crawlers by the previous step, the two lists label as robots all the sessions categorized as Cloud Client, Offline Browser, Link Checker, Feed Fetcher, Library, Validator and Other, and some of the sessions identified as Unknown. 

In the third stage, and in order to label more sessions, we manually label the unique user-agents marked as Unknown from the API. From the total of 2.562 user-agent strings 1.946 are identified as robots. These user-agents are mostly WordPress plugins, reference and citation management tools and custom applications. 

In order to identify humans, we use information from the logs of the publisher's web application. In particular, we label as human all the sessions that come from a logged in user. Sessions previously categorized as Browser by the API, can not be considered humans since robots can mask their user-agent string with one of a known browser \cite{stevanovic2013detection}.

Following the above labeling strategy we managed to label 67.484 sessions, of which 37.922 ($\approx 56\%$) are labeled as robot and 29.562 ($\approx 44\%$) as human. Interestingly, the distribution of the target variable in our dataset is similar to the reported distribution of human and bot traffic on the Web ($\approx 50\%$ each). In addition, we note that our original data remain largely unlabeled, exposing the inability of heuristic algorithms and syntactical log analysis to effectively detect robots~\cite{doran2011web}.

\section{Results}
\label{results}

Here, we contribute empirical results concerning the utility of the proposed semantic features for web robot detection in our real-world case study. We first present and discuss measures of the dependency between each feature (both simple and semantic) and the class variable (human vs bot). Then we compare and contrast simple, semantic and both sets of features in conjunction with a variety of machine learning algorithms for building web robot detection models. 

\subsection{Feature Evaluation}
We discuss the dependency between each feature and the class, as measured by two univariate statistical tests: the F-test in ANOVA and the $\chi^2$ test. Table \ref{table:feature_scores} presents the scores of all features according to these two tests, in descending order.

% \usepackage{booktabs}
% \usepackage[normalem]{ulem}
\begin{table*}[ht]
\centering
\renewcommand{\arraystretch}{1.3}
\caption{The scores of the features in descending order.}
\label{table:feature_scores}
\begin{tabularx}{\textwidth}{@{}XrXr@{}}
\toprule
\multicolumn{2}{c}{F-test}                           & \multicolumn{2}{c}{$\chi^2$ test}                             \\ \midrule
\multicolumn{1}{c}{Feature}     & \multicolumn{1}{c}{Score} & \multicolumn{1}{c}{Feature}     & \multicolumn{1}{c}{Score} \\
\cmidrule(r){1-2} \cmidrule(l){3-4}
\textbf{Boolean Variance}      & 55346.73                  & Session Duration                & 43153243.50               \\
\textbf{Page Similarity}    & 21268.92                  & \textbf{Total Topics}           & 6265437.49                \\
Repeated Requests               & 16867.69                  & \textbf{Unique Topics}    & 2188149.91                \\
\textbf{Page Variance} & 13679.18                  & Total Requests                  & 628569.58                 \\
\textbf{Unique Topics}    & 9695.67                   & Unique Content                  & 310769.30                 \\
Unique Content                  & 5626.69                   & Average Time               & 259439.45                 \\
Session Duration                & 3736.19                   & Standard Deviation Time         & 153834.77                 \\
HTTP Client Error               & 2221.33                   & \textbf{Boolean Page Variance}      & 72611.97                  \\
Total Requests                  & 1593.08                   & \textbf{Page Similarity}    & 2803.16                   \\
\textbf{Total Topics}           & 1584.91                   & \textbf{Page Variance}               & 2208.23                   \\
Average Time               & 909.14                    &  Repeated Requests               & 678.14                    \\
Standard Deviation Time         & 574.04                    & HTTP Client Error & 267.90                    \\
PDF Requests                    & 458.06                    & PDF Requests                    & 83.89                     \\
HTTP Successful                 & 100.65                    & HTTP Successful                 & 25.02                     \\
HTTP Server Error               & 35.54                     & Webservice                      & 22.45                     \\
HTTP Redirection                & 25.27                     & HTTP Server Error               & 18.08                     \\
Webservice                      & 22.49                     & HTTP Redirection                & 9.63                      \\
Multiple Countires              & 3.19                      & Multiple Countires              & 3.18                      \\
\bottomrule
\end{tabularx}
\end{table*}

We first notice that all semantic features are highly ranked by at least one of the two tests. In particular, four out of the five semantic features ({\em Boolean Page Variance}, {\em Page Similarity}, {\em Page Variance} and {\em Unique Topics}) are among the top-5 feature according to the F-test, while two out the five semantic features ({\em Total Topics} and {\em Unique Topics}) are among the top-3 features according to the $\chi^2$ test. These findings are in line with our initial hypothesis that semantic features make a useful representation of sessions for web robot detection in content rich websites.

We also notice that simple features like {\em Repeated Requests}, {\em Session Duration} and {\em Total Requests} are also ranked high by both tests. This is expected, since long sessions with many and repeated requests is typical of the behavior of web robots. 

\begin{figure}[tp]
\centering
\includegraphics[width=\columnwidth]{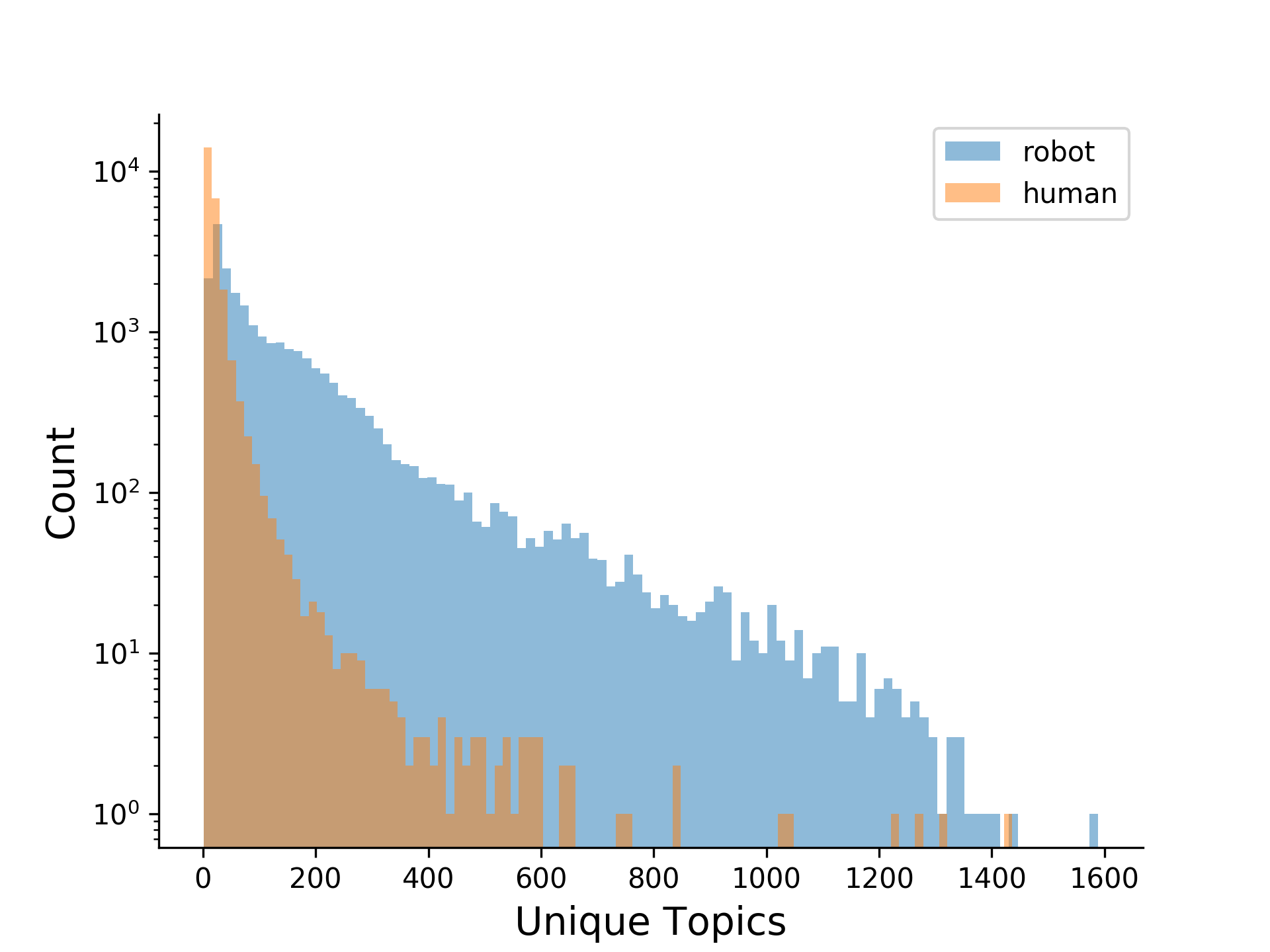}
\caption{A histogram comparing the distribution of human and robots sessions' unique topics in the log scale.}
\label{fig:unique_histogram}
\end{figure}

The {\em Unique Topics} semantic feature is the only feature to be found among the top-5  features according to both tests. Figure \ref{fig:unique_histogram} contrasts the distribution of {\em Unique Topics} in human sessions with that of robot sessions. For this particular graph we randomly under-sample the robot sessions in order to match the number of human sessions ($\approx 29k$). It is evident that robot sessions exhibit a much higher number of {\em Unique Topics} compared to human sessions.  

\subsection{Predictive Modeling}

We split the original training data in two parts: a training set containing 70\% and a test set containing the rest 30\%. The split is done in a time-ordered way, so that the training set contains only sessions that occurred before the test set, in accordance with a real-world deployment. 

We experiment with four different models: a support vector machine with an RBF kernel (RBF), a gradient boosting (GB) model and a multi-layer perceptron (MLP) using scikit-learn~\cite{scikit-learn}, as well as an eXtreme Gradient Boosting (XGB) model\footnote{\url{https://github.com/dmlc/xgboost}}.

Table \ref{table:experiments} presents the F-measure, Balanced Accuracy and G-mean of each model using only the simple features, only the semantic features and, finally, both the simple and the semantic features.

We first see that the best results in all three evaluation measures are achieved by RBF when the semantic features are used by themselves, and by GB when the simple features are used either by themselves or in tandem with the semantic features (in bold typeface). Considering these best results per feature space used, we notice that a decent level of web robot detection accuracy can be achieved using semantic features alone. Simple features lead to better results compared to semantic features when these two types of features are used by themselves. However, the best results in all three evaluation measures are achieved when using both the simple and the semantic features (in underlined typeface). In particular, the F-measure is increased by by 1.06\%, Balanced Accuracy by 1.26\% and G-mean  by 1.34\% compared to using the simple features alone. These findings are evidence that semantic features can lead to improved web robot detection accuracy in content rich websites.    

\begin{table*}[tp]
\renewcommand{\arraystretch}{1.3}
\centering
\caption{Result of the experiments conducted.}
\label{table:experiments}
\begin{tabular}{@{}ccccccccccccc@{}}
\toprule
\multirow{2}{*}{Features} & \multicolumn{4}{c}{F-measure}              & \multicolumn{4}{c}{Balanced Accuracy}      & \multicolumn{4}{c}{G-mean}                 \\ \cmidrule(l){2-5} \cmidrule(l){6-9} \cmidrule(l){10-13} 
                          & RBF    & MLP    & GB              & XGB    & RBF    		& MLP    & GB              & XGB    & RBF    & MLP    & GB              & XGB    \\ \cmidrule(r){1-1}
Simple                    & 0.6552 & 0.7844 & \textbf{0.9075} & 0.905  & 0.6551 		& 0.7685 & \textbf{0.9007}          & 0.898  & 0.5835 & 0.7432 & \textbf{0.8989}          & 0.8961 \\
Semantic                  & \textbf{0.8489} & 0.7497 & 0.8482 & 0.846  & \textbf{0.8484} & 0.7712 & 0.845           & 0.8418 & \textbf{0.8475} & 0.7673 & 0.8432          & 0.8395 \\
Simple \& Semantic        & 0.6484 & 0.8166 & \underline{\textbf{0.9181}} & 0.9177 & 0.6518 & 0.801  & \underline{\textbf{0.9133}} & 0.9127 & 0.5656 & 0.7815 & \underline{\textbf{0.9123}} & 0.9116 \\ \bottomrule
\end{tabular}
\end{table*}

%The above observations comes close to the confirmations of our assumption that information deriving from the content visited can produce valuable semantic characteristics for a web robot detection problem and characterize the (in)consistency of a session.

\begin{figure}[tp]
\centering
\includegraphics[width=\columnwidth]{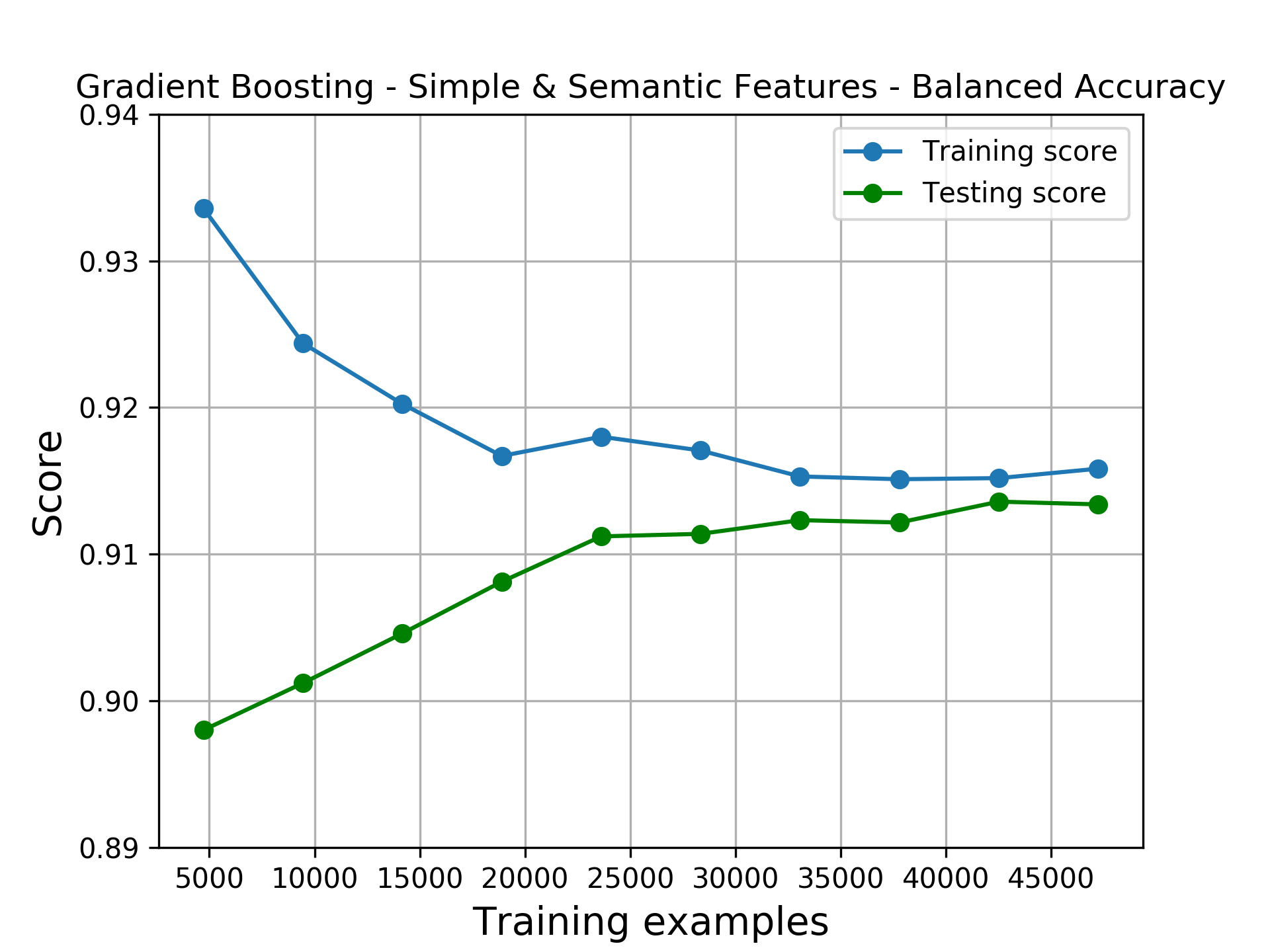}
\caption{Learning curve of the Gradient Boosting algorithm using both the simple and the semantic features.}
\label{fig:curve_all}
\end{figure}

We conclude the discussion of the results with a learning curve plotting the Balanced Accuracy of the best learning algorithm, GB, using the best representation of sessions, both simple and semantic features, for a varying number of training examples (Figure \ref{fig:curve_all}). We see that the training and testing accuracy curves converge when almost half of the available data is used. We can therefore conclude that our model has low variance and there is no need for additional training data to improve the current results. Instead, the complexity of our model should be increased, either by getting additional features or by using polynomial features.     

% Statistical significance... \SL{A searched a lot about this. Paired t-test needs a cross validation to work. I also thought if somehow the probabilities of the scores can help to this. Sth like the entropy of the probabilities. How confident our classifier is, maybe Area Under the Curve? }
% \begin{figure}[tp]
% \centering
% \includegraphics[width=\columnwidth]{figures/lrcurveSimple}
% \caption{Learning curve of the Gradient Boosting algorithm using only the simple features.}
% \label{fig:curve_simple}
% \end{figure}

% \begin{figure}[tp]
% \centering
% \includegraphics[width=\columnwidth]{figures/lrcurveSemantic}
% \caption{Learning curve of the Gradient Boosting algorithm using only the semantic features.}
% \label{fig:curve_semantic}
% \end{figure}

\section{Conclusion \& Future Work}
\label{conclude}
In this work we discuss the use of semantic features on detecting web robots. Based on the assumption that humans typically look for specific information on a particular subject, while on the other hand, most of the web robots go through the content of a website in a uniform fashion, we propose a novel set of semantic features that can potentially increase the discrimination capability of a supervised learning model. We performed an empirical study on real world data, from an academic digital library, by comparing the use of features previous research used and the semantic characteristics we engineered. We focused on the significance of features and we experimented with various learning algorithms. The results of our study seems promising, reaching a balanced accuracy over 90\% when using both semantic and common features while in most of the classifiers tested the use of semantic features increases considerably the classification accuracy. Finally, the scores of the features, deriving by statistical tests, clearly indicate the importance of the semantic features and the consideration need to be taken when selecting the different features. 

% This work discussed our supervised learning approach on detecting web robots in academic publishing. It uses a variety of features deriving from both the log files of a server but also novel characteristics deriving from the content the user visited. These semantic features are crafted by performing a semantic analysis using the LDA algorithm.  in order to support our assumption that the use of semantic features affects positively the prediction model's discrimination capabilities between human and robots. The empirical study, conducted in real-world case, showed promising results ,  

In the future, we plan to make the most out of the semantic analysis by creating more and different features that can better characterize the (in)coherence of a session. Our study showed that the increase of semantic features may augment the discrimination capabilities of a model. Toward this, we are gonna fine tune the parameters of LDA or even use other algorithms like Word2Vec to extract meaningful information from the web pages a user visits.

Our future plans also include the modification of the classes we use for our detection. Following the path of previous research we are going to transform the problem to a multi-class problem, since web robots come with different characteristics and functions. An analysis of the significance of semantic features on the different types of web robots may expose their true intentions while unsupervised techniques may reveal new uncategorized robots.

\bibliographystyle{ACM-Reference-Format}
\bibliography{sample-bibliography} 

\end{document}